# A Statistical Nonparametric Approach of Face Recognition: Combination of Eigenface & Modified k-Means Clustering

**Soumen Bag, Soumen Barik, Prithwiraj Sen and Gautam Sanyal**

Department of Computer Science and Engineering, National Institute of Technology, Durgapur, West Bengal, India

## Abstract

Facial expressions convey non-verbal cues, which play an important role in interpersonal relations. Automatic recognition of human face based on facial expression can be an important component of natural human-machine interface. It may also be used in behavioural science. Although human can recognise the face practically without any effort, but reliable face recognition by machine is a challenge. This paper presents a new approach for recognizing the face of a person considering the expressions of the same human face at different instances of time. This methodology is developed combining Eigenface method for feature extraction and modified k-Means clustering for identification of the human face. This method endowed the face recognition without using the conventional distance measure classifiers. Simulation results show that proposed face recognition using perception of k-Means clustering is useful for face images with different facial expressions.

## 1 INTRODUCTION

Over the past 20 years [1], considerable amount of works have been carried out for automatic recognition of facial expression by different researchers using parametric as well as non-parametric techniques. The work proposed by M.A.Turk & A.P.Pentland [2, 3], in "Eigenfaces for Recognition" the input images is projected onto the face space using principle component analysis called Eigenfaces. The recognition is performed using various distance classifiers and a correct distance-threshold. Sang-Jean Lee et al. [4] developed a system that can locate and track person's head in a complex background and recognization is done using pre-extracted face images on a feature space using PCA. Moon H. et al. [5] designed a generic PCA algorithm for experimenting on FERET image set with varieties of issues such as illumination normalization, effect of compression with JPEG and Wavelet, eigenvectors selection etc. Further, Hyun-chul Kim et al. [6] deals with "face recognition using the mixture-of-eigenfaces method" uses more than one set of eigenfaces from the expectation maximization learning in the PCA mixture model for representation of face images with large variations. The research carried out by Yan Ma et al. [7] stated that during training process average projection vector of different images for each person is used. Recognition uses two thresholds (acceptance and rejection). The work done by M.A. Rabbani et al. [8] used Median instead of mean and with different distance measures in appearance-based statistical eigenface method.

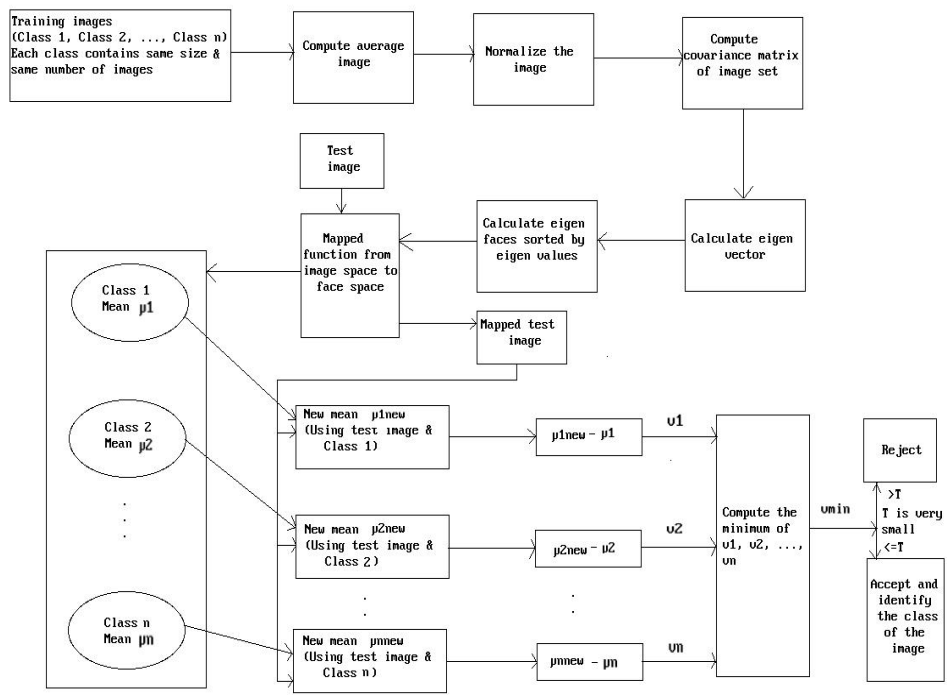

**Fig. 1** Schematic block diagram of the proposed face recognition system

In our proposed work we address a simple yet efficient method for face recognition that considered one of the hurdles of the Eigenface method that uses different distance classifiers. In our travail, deliberating different facial images of a single human face taken together as a cluster, then have applied concept of k-Means clustering for recognition that avoids that exploit a threshold value which is changed under different distance classifiers. This is very sophisticated face recognition algorithm and has comparable performance with conventional eigenface face recognition method.

## 2 MATHEMATICAL FORMULATION

### 2.1 Eigenfaces

- Obtain face images $I_1, I_2, ...I_M$ of same size, where M= No. of images in training image set.
- Transformation of an image ( $I_i \ \varepsilon \ I$ ) to a column vector $\Gamma_i$, where I is the training image
- Mean face:

$$\Psi = 1/M \sum_{i=1}^{M} \Gamma_i \quad (1)$$

where M = No. of images in training image set, $\Psi$= Mean face

- Normalized images:

$$\Phi_i = \Gamma_i - \Psi \quad (2)$$

where $\Phi_i = i^{th}$ normalized training image.
- Covariance matrix:

$$Cov = 1/M \sum_{n=1}^{M} \Phi_n \Phi_n^T = AA^T \quad (3)$$

where Cov = covariance matrix, $A = [\Phi_1 \Phi_2 ..... \Phi_M]$
- Compute the M eigenvector (EV)s (E numbers of EVs are selected based on the descending order of eigenvalues from M) by the equation $u_i = Av_i$, where $v_i$ is the eigenvector of the matrix $A^T A$ and $u_i$ are the eigenvectors of the $AA^T$ matrix (Normalized $u_i$ such that $|u_i|=1$).
- Face space images:

$$\Omega_i = u^T \Phi_i, \quad i = 1 \text{ to } M \quad (4)$$

where $u = [u_1, u_2, ..... u_E]$, $E \leq M$, $\Omega_i$ = face space training images,
- Mean of each class:

$$\pounds_k = 1/P \sum_{i=1}^{P} \Omega_i^k, \quad k = 1 \text{ to } C \quad (5)$$

where $\Omega_i^k = i^{th}$ face space image of $k^{th}$ class, $\pounds_k$ = mean of each $k^{th}$ class, C = no. of classes (each person belongs to a separate class), P = no. of images per class.

## 2.2 Recognition of a Probe Image

- Normalized test image:

$$\Phi_{test} = \Gamma_{test} - \Psi \tag{6}$$

where $\Gamma_{test}$ = vector form of test image, $\Phi_{test}$ = normalized test image

- Mapped test image:

$$\Omega_{test} = u^T \Phi_{test} \tag{7}$$

where $u = [u_1, u_2, \ldots u_E]$, $\Omega_{test}$ = mapped test image in the face space

- Mean with test image:

$$\mu_k = 1/(P+1)\left(\sum_{i=1}^{P} \Omega_i^k + \Omega_{test}\right), k=1 \text{ to } C \tag{8}$$

where $\mu_k$ = mean of each class including test image, $\Omega_i^k$ = represents $i^{th}$ face space image of $k^{th}$ class, C = no. of classes (each person belongs to a separate class), P = no. of images per class.

- Difference of old and new mean:

$$D_k = |£_k - \mu_k|, \quad k = 1 \text{ to } C, \tag{9}$$

where $D_k$ = difference between old and new mean value of $k^{th}$ class

- Minimum mean difference:

$$D_{min} = \text{Min}(D_k), k=1 \text{ to } C \tag{10}$$

where $D_{min}$ = minimum value among all $D_k$

- If $D_{min} <= T$ (where, T is very small)
  => $\Phi_{test} \, \varepsilon \, I$

## 3 PROPOSED COMPUTER ALGORITHM

### 3.1 Feature Extraction Using Eigenfaces

1. Obtain face images of same size and consider that each person belongs to a separate class.

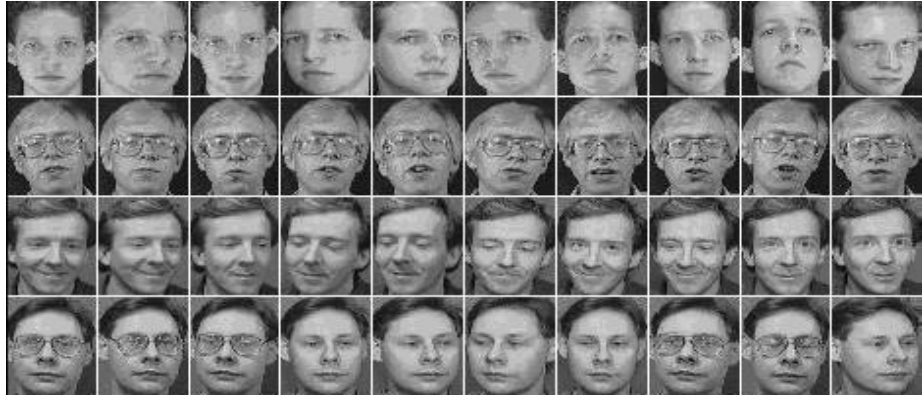

**Fig. 2** Training image set

2. Transform each image into a column vector.
3. Compute the average face vector (Eq. 1).
4. Subtract the mean face from all of the training images (Eq. 2) for normalization (empirical mean distance is zero).
5. Construct the Covariance Matrix of the training image set (Eq. 3).
6. Compute the eigenvectors of the covariance matrix using the characteristic equation of matrix. These eigenvectors are called eigenfaces.
7. Among the eigenvectors (Obtained form step 6) compute the spectral density of the image set by selecting the E (E<=M) number of eigenvectors which have positive eigenvalues and organized as descending order.
8. Mapping the training images from image Space to face Space using the eigenvectors obtain from step 7 (Eq. 4).
9. Computing the Mean of Each Class in the training image set (Eq. 5).

### 3.2   Face Recognition Using Modified k-Means Clustering

1. Normalize the test image by subtracting the mean from it (Eq. 6).
2. Project the normalized test image into eigenspace (Eq. 7).
3. Calculation of mean of each class including the test image within each class (Eq. 8).
4. Find out the differences between previous mean (without probe image) and new mean (with probe image) for each class (Eq. 9).
5. Computing the minimum value (MIN) among all mean differences computed in step 4 (Eq. 10).
6. If "MIN" is less than or equal to the threshold (threshold tending to zero), then it is a known image, hence display the class which has minimum difference value. Otherwise it is an unknown image.

## 4  EXPERIMENTAL RESULT

In our experimental purpose, standard ORL face database is used which contains a set of 40 people with 10 images (representing different facial expressions) of each face. For all the cases a subset of images are used as training set and the rest of the images as testing set.

### 4.1  Case Study 1

Table 1 represents the result of average percentage of success when we have used one image per person of the training image set.

**Table 1**   Represents the result of average percentage of successes

| NITDS | NII | EKRM(%) |
|-------|-----|---------|
| 8     | 40  | 87.5    |
| 16    | 55  | 82.1    |
| 20    | 80  | 79.4    |
| 26    | 120 | 75.7    |
| 32    | 150 | 72.1    |

**NITDS:**  No. of images in the training data set, **NII:** No. of input images,
**EKRM:**  Eigen_k-Means Recognition Method

### 4.2  Case Study 2

Table 2 represents the average percentage of success when we have used more than one image (2, 3, 4, 5 and 6) of each person (considering 32 persons) of the training image set.

**Table 2**   Represents the result of average percentage of successes

| NIPP | NII | EKRM(%) |
|------|-----|---------|
| 2    | 50  | 72.0    |
| 3    | 70  | 74.6    |
| 4    | 90  | 78.0    |
| 5    | 120 | 80.2    |
| 6    | 150 | 84.0    |

**NIPP:** No. of images per person in the training set, **NII:** No. of input images,
**EKRM:** Eigen_k-Means Recognition Method

## 5 CONCLUSION

Under this new methodology we can avoid the use of various distance measurement technique and computation of correct threshold value which is the core parameter of existing eigenface method. Very little work has been done on avoiding the use of distance classifiers and threshold measurements in Eigenface based method. Previous work [9] has given emphasis on selecting the correct distance measure classifiers. When we apply our new methodology for both the case (Case Study 1 & 2), the success rates are satisfactory. This method provides a simple and fast solution while maintaining standard prediction accuracy.

## References

1. R. Chellappa, C. Wilson, S. Sirohey, "Human and Machine Recognition of Faces: A Survey," Proc. IEEE, vol. 83, no. 5, pp. 705–741, 1995.
2. M. Turk and A. Pentland, "Eigenfaces for Recognition", Journal of Cognitive Neuroscience, vol 3, no. 1, pp. 71–86, 1991.
3. M. A. Turk and A.P. Pentland, "Face recognition using eigenfaces". In Proc. of the CVPR, pp. 586–591, June 1991.
4. S. J. Lee, S. B. Jung, J. W. Kwon, S. H. Hong, "Face Detection and Recognition Using PCA", In Proc. of the IEEE TENCON, vol. 1, pp. 84–87, South Korea, 1999.
5. H. Moon, P. J. Phillips, "Computational and performance aspects of PCA-based face-recognition algorithms", Perception, vol. 30, no. 3, pp. 303 – 321, 2001.
6. H. C. Kim, D. Kim, S. Y. Bang, "Face recognition using the mixture-of-eigenfaces method", Pattern Recognition Letters, vol. 23, no. 13, pp. 1549–1558, 2002.
7. Y. Ma, S. B. Li, "The Modified Eigenface method using Two Thresholds", International journal of Signal Processing, vol. 2, no. 4, pp. 236–239, 2006.
8. M. A. Rabbani, C. Chellappan, "A Different Approach to Appearance–based Statistical Method for Face Recognition Using Median ", IJCSNS, vol. 7, no. 4, pp. 262–267, 2007.
9. W. S. Yambor, B. A. Draper, J. R. Beveridge, "Analyzing PCA-based Face Recognition Algorithm: Eigenvector Selection and Distance Measures", Computer Science Department, Colorado State University, Fort Collins, CO, USA 80523, July 2000.